\documentclass{article}
\newcommand{\bst}{\mathbf{s}_{t}}

\newcommand{\bat}{\mathbf{a}_{t}}
\newcommand{\bss}{\mathbf{s}_{t+1}}
\newcommand{\bso}{\mathbf{s}_{0}}

\usepackage[final]{cpal_2024}
\usepackage{amsmath,amsfonts,amssymb,graphicx}
\usepackage{amsmath, bm}
\usepackage{graphicx}
\usepackage{textcomp}

\usepackage{enumitem}
\usepackage{subcaption}
\usepackage{algpseudocode}
\usepackage{relsize}
\usepackage[export]{adjustbox}
\usepackage{array,multirow}
\usepackage{multirow}
\usepackage{lscape}
\usepackage{bbm}
\usepackage{graphicx}
\usepackage{booktabs}
\usepackage{url}
\usepackage{xcolor}         
\usepackage{tcolorbox}
\usepackage{pgfplots}
\usepackage{amsmath}
\usepackage{amssymb}
\usepackage{dsfont}
\usepackage{mathtools}
\usepackage{amsthm}
\usepackage{enumitem}
\usepackage{wrapfig}
\usepackage{algorithm}
\usepackage{caption,subcaption}
\usepackage{hyperref}
\usepackage{url}
\usepackage{enumitem}
\usepackage{amsmath}
\usepackage{amsfonts}
\usepackage{amssymb}
\usepackage{xcolor}
\usepackage{subcaption}
\usepackage{algorithm,algorithmicx}
\usepackage{relsize}
\usepackage[export]{adjustbox}
\usepackage{array,multirow,graphicx}
\usepackage{multirow}
\usepackage{graphicx}
\usepackage{lscape}
\usepackage{amsthm}
\usepackage{bbm}
\usepackage{graphicx}
\usepackage{booktabs}
\usepackage{wrapfig}
\usepackage{url}
\usepackage{tikz}
\definecolor{lightblue}{rgb}{0.678, 0.847, 0.902} 
\definecolor{lightyellow}{rgb}{1.0, 1.0, 0.8}
\usepackage{tcolorbox}

\newtheorem{theorem}{Theorem}

\usepackage{url}

\newtheorem{definition}{Definition}

\newtcolorbox[auto counter, number within=section]{mybox}[2][]{colback=#2!5!white,colframe=#2!75!black,
fonttitle=\bfseries, title=Box~\thetcbcounter: #1,#2,#2}

\title{A Dataless Reinforcement Learning Approach to Rounding Hyperplane Optimization for Max-Cut}

\author{
  Gabriel Maliakal\textsuperscript{1}, ~Ismail Alkhouri\textsuperscript{1,2}, \\ ~Alvaro Velasquez\textsuperscript{3}, ~Adam M Alessio\textsuperscript{1}, ~Saiprasad Ravishankar\textsuperscript{1} \\ 
  \textsuperscript{1}Michigan State University\\
  \textsuperscript{2}University of Michigan, Ann Arbor \\
  \textsuperscript{3}University of Colorado, Boulder \\
}

\begin{document}

\maketitle

\begin{abstract}
The Maximum Cut (MaxCut) problem is NP-Complete, and obtaining its optimal solution is NP-hard in the worst case. As a result, heuristic-based algorithms are commonly used, though their design often requires significant domain expertise. More recently, learning-based methods trained on large (un)labeled datasets have been proposed; however, these approaches often struggle with generalizability and scalability. A well-known approximation algorithm for MaxCut is the Goemans-Williamson (GW) algorithm, which relaxes the Quadratic Unconstrained Binary Optimization (QUBO) formulation into a semidefinite program (SDP). The GW algorithm then applies hyperplane rounding by uniformly sampling a random hyperplane to convert the SDP solution into binary node assignments. In this paper, we propose a training-data-free approach based on a non-episodic reinforcement learning formulation, in which an agent learns to select improved rounding hyperplanes that yield better cuts than those produced by the GW algorithm. By optimizing over a Markov Decision Process (MDP), our method consistently achieves better cuts across large-scale graphs with varying densities and degree distributions.

\end{abstract}


\section{Introduction}
The Maximum Cut (MaxCut) problem is one of Karp's classical NP-complete problems~\cite{karp2009reducibility}. Given a graph $G = (V, E)$, where $V$ is the set of vertices and $E$ is the set of edges, the MaxCut problem seeks a subset of nodes $S \subset V$ such that the number of edges between $S$ and its complement is maximized. MaxCut has numerous applications across domains such as image segmentation~\cite{MaxCut_segment, maxcut_segmentation_cardiac}, VLSI design~\cite{MaxCut_vlsi_barahona1988application}, and network robustness~\cite{maxcut_network_xi2007method}.

Several approaches have been proposed to tackle MaxCut, including exact solvers based on Integer Linear Programming (ILP)—which employ Branch-and-Bound methods~\cite{MaxCut_bnb_lu2021}—and Quadratic Unconstrained Binary Optimization (QUBO). However, due to their reliance on integer variables, ILP and QUBO formulations typically scale poorly, often requiring prohibitively long computation times on large and/or dense graphs. To address these scalability issues, approximate methods have been introduced. For instance, the QUBO formulation of MaxCut can be relaxed into a semidefinite program (SDP). While SDPs are convex and solvable in polynomial time, their solutions require rounding techniques to recover a valid cut from the continuous solution. A notable example is the Goemans-Williamson (GW) algorithm~\cite{goemans1995improved}, which offers a well-known approximation guarantee. Heuristic-based approaches such as Breakout Local Search (BLS)~\cite{benlic2013breakout} have also been developed, though they lack theoretical guarantees and often require domain-specific tuning.

More recently, learning-based and data-centric methods have been proposed for MaxCut and other combinatorial optimization problems~\cite{s2v-dqn, sutton2018reinforcement, eco-dqn, pCQO}. These include supervised, unsupervised, and reinforcement learning (RL) approaches. However, such methods often require a large number of (un)labeled training graphs and may suffer from limited generalization and poor scalability to out-of-distribution instances.

Motivated by (\textit{i}) the desire to eliminate reliance on training data, and (\textit{ii}) the goal of surpassing the GW bound, we propose an RL-based solver that operates on the SDP solution. Our method uses a fully connected, multi-head neural network and does not depend on heuristics or training datasets. We compare our method to a GPU-accelerated GW implementation that employs parallel hyperplane sampling for rounding, which we refer to as parallelized GW (pGW).

\paragraph{Contributions:} We introduce a self-supervised, non-episodic reinforcement learning algorithm that operates without any training graphs. Our method learns to generate hyperplanes that maximize the cut value by modeling a Gaussian distribution over hyperplanes, initialized from a uniform distribution and guided by the SDP solution from the Goemans-Williamson (GW) formulation. This Gaussian is parameterized by a fully connected, multi-head neural network with a value head. The learning process is framed as a Markov Decision Process (MDP), where new hyperplanes are sampled based on the current policy and used to round the SDP solution. An Actor-Critic network is trained using Proximal Policy Optimization (PPO)~\cite{schulman2017proximal} to improve the policy. We demonstrate that our method consistently outperforms the parallelized Goemans-Williamson baseline (pGW) on large graphs from the Gset dataset as well as on synthetic Erdős-Rényi (ER) graphs across different densities.

\section{Background \& Motivation}


Let an unweighted and undirected graph be defined as $G(V,E)$, where $V=\{1,...,n\}$ is the set of nodes and $E \subset V\times V$ (with $|E|=m$) is the set of edges that consists of pairs of vertices that are connected by an edge. The Maximum Cut (MaxCut) problem is formally defined next. 
\begin{definition}[Maximum Cut (MaxCut)] 
Given an undirected and unweighted graph $G=(V,E)$, the MaxCut problem aims to partition $V$ into two disjoint subsets $S$ and $S^c = V\setminus S$ such that the total number of edges crossing the cut (i.e., edges with one end in $S$ and the other in $S^c$) is maximized.
\end{definition}
Define a vector $\mathbf{x}\in \{0,1\}^n$ such that every entry $x_i$ corresponds to a node $i\in V$, and vector $\mathbf{y}\in \{0,1\}^m$ is such that every entry $y_{i,j}$ corresponds to an edge $(i,j)\in E$ where we consider each tuple $(i,j)$ to be an index for each element in $y$. Then, the ILP of the MaxCut problem is given as 
\begin{align*}\label{eqn: cut ILP}
    \max_{\mathbf{x}\in\{0,1\}^{n},\mathbf{y}\in\{0,1\}^{m}} \sum_{(i,j)\in E}& y_{i,j} \quad 
    \text{s.t.} \quad y_{i,j}\leq x_{i}+x_{j}, \quad y_{i,j}\leq 2-x_{i}-x_{j},\quad\forall (i,j)\in E\:. \tag{\texttt{ILP}}
\end{align*}
In the \eqref{eqn: cut ILP}, there are $n+m$ binary variables. An alternative integer problem, which has only $n$ variables, is the following MaxCut Quadratic Unconstrained Binary Optimization (QUBO). 
\begin{align*}\label{eqn: cut qubo}
    \max_{\mathbf{x}\in \{-1,1\}^{n}} \frac{1}{2}\sum_{(i,j)\in E}(1-x_{i}x_{j})\:.  \tag{\texttt{QUBO}}
\end{align*}
%

We note that the cut value in graph $G$ of some binary vector $\mathbf{x} \in \{-1,1\}^n$ is given by the objective function in \eqref{eqn: cut qubo}. As both of the above optimization problems are binary programs, they do not scale well as $n$ and $m$ increase. Therefore, alternative approaches have been explored such as the MaxCut SDP relaxation we describe next. 

\subsection{MaxCut SDP \& The GW Algorithm}

The pioneering work in \cite{goemans1995improved} introduced the MaxCut SDP formulation for which they first define each node $i\in V$ by a multi-dimensional unit norm vector, $\mathbf{x}_{i} \in \mathbb{S}^{n-1}$, where $\mathbb{S}^{n-1}$ denotes the surface of the $n$-dimensional unit sphere. Then, the MaxCut SDP optimization problem is
%
%
%
%
\begin{align*}\label{eqn: cut sdp}
    \max_{\mathbf{x}_i\in\mathbb{S}^{n-1},\forall i \in V} \frac{1}{2}\sum_{(i,j)\in E}(1-\mathbf{x}_{i}\cdot \mathbf{x}_{j})\:, \tag{\texttt{SDP}}
\end{align*}
where $\mathbf{x}_{i}\cdot \mathbf{x}_{j}$ denotes the dot product between the unit-norm vectors $\mathbf{x}_{i}$ and $\mathbf{x}_{j}$.
Although SDPs are convex optimization programs and can be solved in polynomial time, the solution of \eqref{eqn: cut sdp}, which we denote by $\mathbf{x}_i^*, \forall i\in V$, do not directly correspond to a MaxCut solution on the graph\footnote{We note that \eqref{eqn: cut sdp} is convex, but not strongly convex. This means that while there may exist a unique minimum objective value, there could exist multiple minimizers.}. Therefore, rounding techniques (or approximation methods) are used to map $\mathbf{x}_i^*$ to actual MaxCut solutions. 

\begin{figure}
    \centering
    \includegraphics[width=1.0\linewidth]{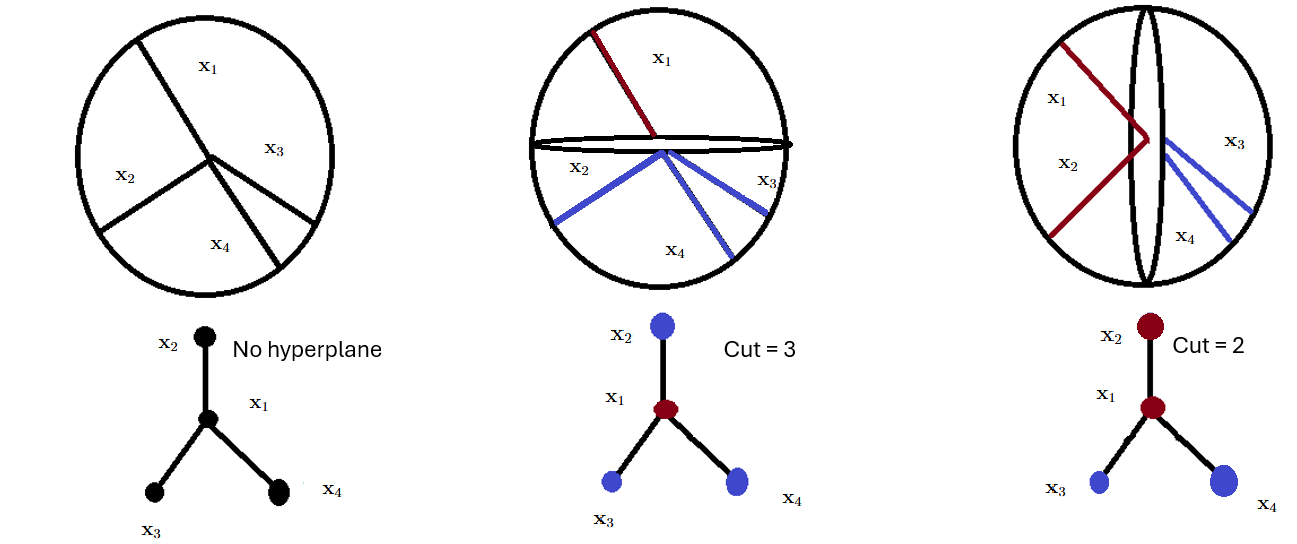}
    \caption{Figure shows the effect of different rounding planes on node assignments. The SDP formulation in this example using the GW method yields solution vectors in $\mathbb{S}^{n-1}$. These are shown as vectors $\mathbf{x}_{1},\mathbf{x}_{2},\mathbf{x}_{3}$,   and $\mathbf{x}_{4}$ on the circle. Different choices of rounding planes assign nodes differently leading to different values of cut.} 
    \label{fig:rounding}
\end{figure}

The most notable approximation algorithm is the Goemans-Williamson (GW) Algorithm \cite{goemans1995improved} (also known as random sampling in Fig\ref{fig:Gset43_main} and Fig\ref{fig:ER200_1000_main}). The GW algorithm procedure consists of first obtaining a random hyperplane through the origin whose unit normal vector is sampled from the uniform distribution over the unit sphere. This normal is denoted as $\mathbf{r}$ with $\mathbf{r} \sim \mathbb{U}(\mathbb{S}^{n-1})$. The hyperplane is then used to partition the indices of the vectors from the SDP solution, i.e., $i$ in $\mathbf{x}^*_i$, into the disjoint sets $S$ and $S^c$. This partitioning takes place by examining the sign of the dot product between $\mathbf{r}$ and $\mathbf{x}_i$. This is shown in Fig. \ref{fig:rounding}. 

Given $\mathbf{r}$, formally, the final solution is obtained as the set $S = \{i\in V \mid \mathbf{x}_{i}^{*}\cdot \mathbf{r} \geq 0\}$. This random sampling procedure (GW algorithm) is further supported by the following theorem. 




\begin{theorem}[Optimality \cite{goemans1995improved}]\label{th: GW}
If $Z(\mathbf{r})$ is the cut value produced by a single random sampling vector $\mathbf{r}$, $Z^{*}_{Q}$ is the optimal value of the objective function in \eqref{eqn: cut sdp}, and $Z^{*}_{P}$ is the optimal value of \eqref{eqn: cut qubo}, then, with $\alpha \approx 0.878$, the GW algorithm guarantees that
\begin{align*}
    \mathbb{E}_{\mathbf{r} \sim \mathbb{U}(\mathbb{S}^{n-1})}\big[Z(\mathbf{r})\big] \geq \alpha Z^{*}_{Q} \geq \alpha Z^{*}_{P}\:.
\end{align*} 
\end{theorem}
Theorem~\ref{th: GW} shows that the GW algorithm guarantees that, in expectation, uniformly sampling a random hyperplane and applying it to the SDP solution yields a cut with value at least 0.878 times the optimal MaxCut value.

\begin{minipage}[t]{0.48\textwidth}
\hrule
\vspace{0pt}
\captionof{algorithm}
{Parallelized GW (pGW)}
\hrule
\begin{algorithmic}[1]
\Require Graph $G$, \# of samples $B$, $Inc=\{\}$
\Ensure Average and incumbent Cut values $avg\_cut$ and $Cut(Inc)$
\State Obtain $\mathbf{x}^*_i, \forall i\in V$ from \eqref{eqn: cut sdp}
\State Initialize Average Cut avg\_cut = 0
\For{$B$ iterations}
    \State $\mathbf{r} \sim \mathbb{U}(\mathbb{S}^{n-1})$
    \State $S = \{i\in V \mid \mathbf{x}_{i}^{*}\cdot \mathbf{r} \geq 0\}$
    \State $avg\_cut = avg\_cut  + \frac{Cut(S)}{B} $
    \If{$Cut(S)>Cut(Inc)$}
        \State $Inc \leftarrow S$
        \State $max\_cut \leftarrow Cut(S)$
        \EndIf
        \EndFor
\end{algorithmic}
\hrule
\label{alg:pgw}
\end{minipage}
\hfill
\begin{minipage}[t]{0.48\textwidth}
\vspace{0pt}

Inspired by recent advances in parallel computing, in this paper, we consider a GPU-accelerated version of the GW algorithm where we sample a batch of hyperplanes in parallel and report both the average and maximum cut values for comparison. We will term this algorithm as parallelized GW (pGW) and use it as the main baselines in our paper. Algorithm~\ref{alg:pgw} presents the procedure. Initialized with graph $G$, number of samples $B$, and $Inc$ (an empty set for saving the incumbent which is the best solution thus far), $B$ samples are drawn in parallel and the algorithm returns the set and value of the largest cut. 
\end{minipage}





Motivated by departing from the need of training datasets, given the SDP solution, in this paper, we explore the possibility of using RL to learn a neural-network-parameterized non-uniform distribution from previously sampled hyper-planes such that, in expectation, we obtain better cuts than pGW. This distribution is specified by a Gaussian distribution with mean and covariance that are conditioned on previous choices of rounding hyperplanes with the initial hyperplane sampled from a uniform distribution over the unit sphere.

\section{Proposed Method} 

In this section, we present our method. We begin by formulating the MaxCut problem as a Markov Decision Process (MDP), followed by a description of the neural network–parameterized Gaussian distribution and its connection to the MDP policy. We then discuss how the agent is trained in a data-less setting using infinite-horizon reinforcement learning with PPO. Finally, we provide a theoretical justification for the proposed approach. 


\subsection{The Formulation of the Markov Decision Process for MaxCut}

We start by formally defining the state and action spaces, and then the transition function. 
\begin{definition}[State space]
The state at time $t$ is defined as $\bst \in \mathcal{S} \subseteq \mathbb{S}^{n-1}$, where the initial state is $\bso \sim \mathbb{U}(\mathbb{S}^{n-1})$. The later states are normalized actions sampled from policy given the current state, defined as $\bss=\frac{\bat}{||\bat||}$. 
\end{definition}
Given a state $\bst$, graph $G(V,E)$, and the SDP solution from \eqref{eqn: cut sdp}, we use
\[\label{eqn: cut}
\text{Cut}(\mathbf{s}) := \sum_{(i,j) \in E} \frac{1}{2} \left(1 - \text{sgn}\left[(\mathbf{x}_i \cdot \mathbf{s})(\mathbf{x}_j \cdot \mathbf{s})\right]\right)\:,
\]
to denote the cut value
where $\text{sgn}[\cdot]$ is the signum function and $\mathbf{s} \in \mathbb{S}^{n-1}$ is the normal of a  new rounding hyperplane and $\mathbf{x}_{i}$ is the i-th column of the SDP solution $\mathbf{X}^{*}$. 
\begin{definition}[Action space]
The action at time $t$ is defined as $\bat \in \mathcal{A} \subseteq \mathbb{R}^{n}$, and is obtained by sampling from the neural network-parameterized Gaussian distribution that we define in Definition~\ref{def: policy}. 
\end{definition}
\begin{definition}[State transition function] The next state is obtained using
$T: \mathcal{A} \times \mathcal{S} \rightarrow \mathcal{S}$ that produces 
\[
T(\bat, \bst) = \frac{\bat}{\|\bat\|} = \bss
\]
\end{definition} The transition to next state is deterministic given an action.

Next, we define the reward signal, full MDP, and the NN-parameterized policy. 
\begin{definition}[Reward]
The reward in our MDP is defined as the difference in cut values after choosing an action, i.e., 
\[
R(\bst, \bat) = \text{Cut}\left(\bss \right) - \text{Cut}(\bst)\:.
\label{eqn:reward}\]
\end{definition}
\begin{definition}[Agent MDP] 
The MDP is defined as a tuple given by a state, an action, reward function, and corresponding next state, $\mathcal{M}=(\bst,\bat,R(\bst,\bat),\bss)$.
\end{definition} 
\begin{definition}[Policy]\label{def: policy}
Actions are sampled from a policy modeled as a parameterized Gaussian distribution:
\[
\pi_{\theta}(\bat \mid \bst) = \mathcal{N}\left(\boldsymbol{\mu}_{\theta}(\bst), diag(\boldsymbol{\Sigma}_{\theta}(\bst)\right))\:,
\] where $\boldsymbol{\mu}_{\theta}(\bst)$ is the mean vector and $\textrm{diag}(\boldsymbol{\Sigma}_{\theta}(\bst))$ is the diagonal covariance matrix, where the entries of both are set to be the output of a policy network, parameterized by $\theta$ that takes state $\bst$ as input. The action $\bat$ is sampled from the policy, i.e., \(\bat \sim \mathcal{N}(\boldsymbol{\mu}_{\theta}(\bst)),diag(\boldsymbol{\Sigma}_{\theta}(\bst)))\).
\end{definition}
%

\begin{definition}[Trajectory ($\tau$)]
A list of state-action tuples for every time step in an episode of length T time steps i.e., $\tau = \{(\bss,\bst, \bat)\}_{t=0}^{T}$.
\end{definition}



In this paper, we employ an infinite-horizon setting, where the trajectory never ends. Empirically, we observe that this takes less time to reach favorable cut values. In general, the goal of infinite-horizon RL is to maximize the expected reward obtained in the MDP. Given some policy $\pi$, this is defined as in \cite{sutton2018reinforcement} by
\begin{align*}
    r(\pi) &= \lim_{t\rightarrow\infty}\mathbb{E}\big[R(\mathbf{s}_{t},\mathbf{a}_{t})|\mathbf{s}_{0},\mathbf{a}_{0:t-1} \sim \pi\big]\:,
\end{align*}
where $\mathbf{a}_{0:t-1}$ are actions sampled from time steps $0$ to $t-1$ and $\mathbf{s}_{0}$ is the initial state.
\begin{definition}[Value]
     Given a state $\mathbf{s}$, the value function is defined as the expected sum of differential rewards given that the trajectory starts from state $\mathbf{S}_{0} = \mathbf{s}$. \[V(\mathbf{s})=\mathbb{E}_{\bat,\bss,... \sim \pi_{\theta}}\big[\sum_{t=0}^{\infty}(R(\bat,\bst)-r(\pi))|\mathbf{S}_{0}=\mathbf{s}\big]\:.\]
\end{definition}
In our paper, $V(\mathbf{s})$ is parameterized by a neural network with parameters $\phi$. Therefore, we denote it by $V_{\phi}(\mathbf{s})$.
%

In our formulation, both the state and action spaces are continuous and multidimensional. Previous works have approached this setting with Deep Q Networks~\cite{mnih2015human}, Deterministic Policy Gradient (DPG)~\cite{ddpg}, or Proximal Policy Optimization (PPO)~\cite{schulman2017proximal}. We employ PPO because it gives us the ability to parameterize both the policy and the value functions for continuous actions and states~\cite{sutton2018reinforcement}. The agent used in our paper is a parameterized Gaussian distribution whose mean and covariance are obtained by a neural network that we describe in the next subsection. 

In our framework, RL guides the GW approach by treating the selection of the hyperplane as the action space and the difference in cuts as the reward. In doing so, we treat the SDP solution as the environment. This contrasts other methods in literature such as \cite{defer_ahn2020learning} which uses conventional RL over multiple environments, i.e., multiple different instances of MaxCut. The agent starts with no data; however, by interacting with the environment it generates transitions which are used in the PPO algorithm to adjust weights of the agent. In our case, the environment is the structure of the graph itself after it has been approximately solved.

\subsection{Neural-Network-based RL Agent}

Previous ML-based MaxCut solvers, such as~\cite{s2v-dqn} and~\cite{eco-dqn}, utilize Graph Neural Networks (GNNs). In contrast, although our method also relies on graph structure through the SDP solution, we employ a fully connected neural network with ReLU and Softplus activations to model the value head and the parameterized Gaussian distribution. Specifically, the network includes two additional heads: one for the mean vector and one for the covariance matrix.

Importantly, we do not incorporate the graph structure directly into the architecture of the neural network. Instead, the graph is used solely for computing the reward function, as defined in Definition~\eqref{eqn:reward}. Our aim is to find an optimal rounding hyperplane that maps the columns of the SDP solution to binary assignments. Using a fully connected network to parameterize the Gaussian policy simplifies the architecture and reduces the number of learnable parameters.







\begin{figure}
    \centering
    \includegraphics[width=1\linewidth]{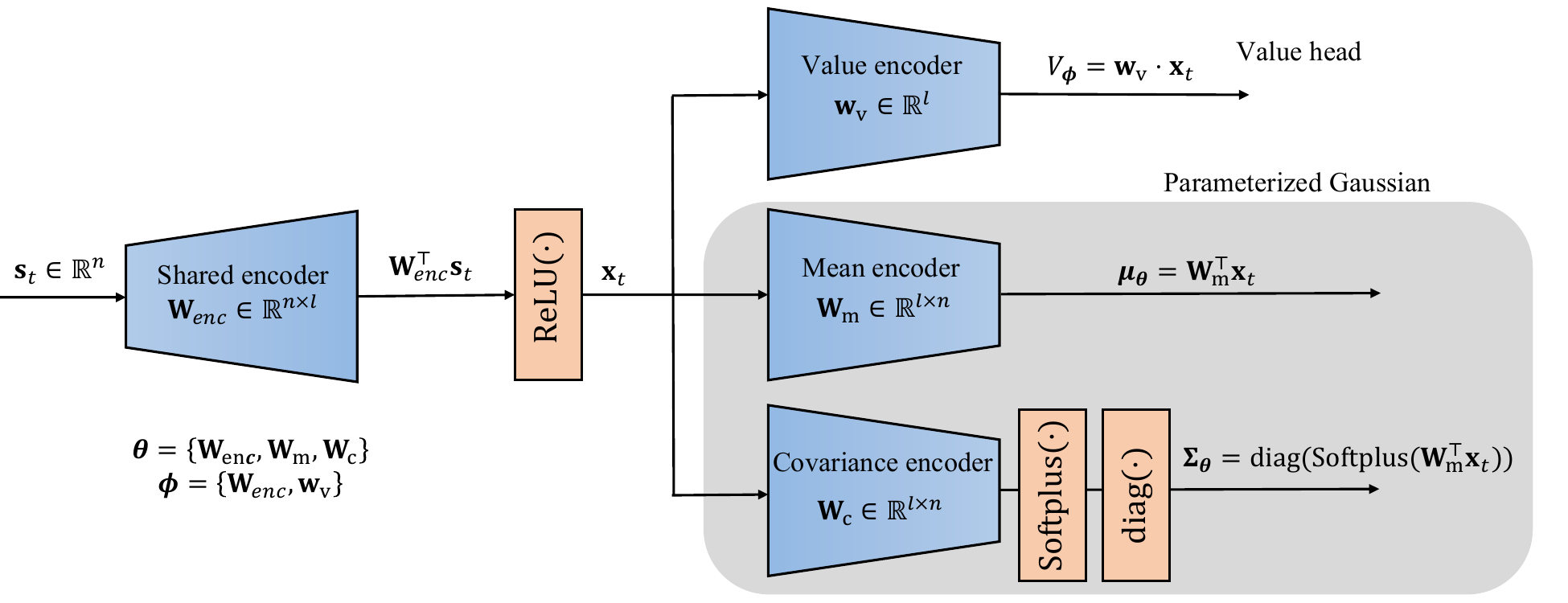}
    \caption{The architecture of the actor-critic, fully connected, multi-head neural network used in our approach. The state $\mathbf{s}_t$ is first passed through a shared encoder, $\mathbf{W}_{\textrm{enc}}$. The encoder output, denoted by $\mathbf{x}_t$ after a ReLU activation, is then fed in parallel into three heads: a value head that outputs $V_{\boldsymbol{\phi}}$, and two heads that produce $\boldsymbol{\mu}_{\boldsymbol{\theta}}$ and $\boldsymbol{\Sigma}_{\boldsymbol{\theta}}$, which parameterize the Gaussian distribution.}
    \label{fig: NN_archi}
\end{figure}



Our agent uses a fully connected network with shared encoder connected to three output heads. The input to the network is the current state $\bst \in \mathbb{S}^{n-1}$, the normal of the rounding hyperplane. Define $\mathbf{W}_\textrm{enc} \in \mathbb{R}^{n\times l}$ to be the shared encoder weight matrix, where $l$ is the size of the hidden dimension. The shared encoder is followed by a ReLU activation function. 

There are three parallel output heads connected to the shared encoder. Define $\mathbf{w}_\textrm{v} \in \mathbb{R}^l$, to be weights of the fully connected layer in the value head. Let $\mathbf{W}_\textrm{m}, \mathbf{W}_\textrm{c} \in \mathbb{R}^{l\times n}$ be the weights of the mean and covariance heads, respectively. Let $\mathbf{x}_t:=\text{ReLU}(\mathbf{W}^{\top}_\textrm{enc}\mathbf{s}_t)$, then the outputs are: 
\begin{equation}\label{eqn: value head}
V_{\boldsymbol{\phi}}(\mathbf{s}_t) = \mathbf{w}_\textrm{v} \cdot \mathbf{x}_t\:,
\end{equation}
\begin{equation}\label{eqn: mean head}
\boldsymbol{\mu}_{\boldsymbol{\theta}}(\mathbf{s}_t) = \mathbf{W}^{\top}_\textrm{m}  \mathbf{x}_t\:,
\end{equation}
\begin{equation}\label{eqn: cov head}
\boldsymbol{\Sigma}_{\boldsymbol{\theta}}(\mathbf{s}_t) = \text{diag}(\text{Softplus}(\mathbf{W}^{\top}_\textrm{c}  \mathbf{x}_t))\:,
\end{equation}
where $\boldsymbol{\phi} = \{\mathbf{W}_\textrm{enc}, \mathbf{w}_\textrm{v}\}$ and $\boldsymbol{\theta} = \{\mathbf{W}_\textrm{enc}, \mathbf{W}_\textrm{c}, \mathbf{W}_\textrm{m}\}$ are used for notational simplicity. In \eqref{eqn: cov head}, $\text{diag}(\cdot)$ converts a vector into a diagonal matrix, and $\text{Softplus}(\cdot) = \log{(1+\exp{(\cdot)})}$ denotes the softplus element-wise activation function used to ensure that there are not any non-positive values in the parameterized covariance matrix. 

These mean and covariance outputs represent the Gaussian policy. The third output returns the value of input state. An action $\mathbf{a}_{t}$ is sampled from this Gaussian distribution and then normalized to a unit vector by dividing by the $\ell_2$ norm of the output to produce the next state $\bss$. See Figure\ref{fig: NN_archi} for a block diagram of our neural network.

\subsection{Training the agent}

The agent is trained using PPO\cite{schulman2017proximal}. The policy head which consists of the mean and covariance outputs for the parameterized Gaussian distribution is trained using PPO.
The agent interacts with the environment to produce transitions from current state to next state by sampling an action from the policy. Along with $\bat,\bst,\bss$, the target for the value network $V_{t} = R(\bst,\bat) + \gamma V_{\phi_{old}}(\bss)$, which is the one-step Temporal Difference (TD) update used as target to train the Value Network\cite{sutton2018reinforcement}, $\pi_{\theta_{old}}(\bat|\bst)$ probability of action given state and advantage which is estimated as $\hat{\delta}_{t,\pi_{old}} = R(\bst,\bat)+\gamma V_{\phi_{old}}(\bss)-V_{\phi_{old}}(\bst)$ are recorded in the buffer. \begin{align}\label{eqn:l_ppo}
     \; L_{PPO}(\theta) = \; \mathbb{E}_{t} [
        \min ( 
            & ratio_{t}(\theta) . \hat{\delta}_{t,\pi_{old}},  
             \text{clip}\left( ratio_{t}(\theta),\ 1 - \epsilon,\ 1 + \epsilon \right) . \hat{\delta}_{t,\pi_{old}}
        )
    ] \:,
\end{align}
\begin{equation}
    L_{VF}(\phi) = [V_{t}-V(\bst,\phi)]^{2}. \label{eqn:l_vf}
\end{equation}
Since the encoder is shared, we train the value and policy head simultaneously by doing stochastic gradient descent on the combined objective function: \begin{equation}
    \min_{\theta,\phi}L_{agent}(\phi,\theta) = -L_{PPO}(\theta)+L_{VF}(\phi) \label{eqn:l_agent}
\end{equation} To perform the optimization, a batch of  transitions $(\bst, \bat, \bss, V_{t}, \pi_{\theta_{old}}(\bat|\bst),\hat{\delta}_{t,\pi_{old}})$ are sampled from the buffer (given by $Buf$ in algorithm \ref{alg:gen_trajectory_update_agent}).

To train the value and policy heads, \ref{eqn:l_agent} is minimized.
Here $ratio_{t}$ is the ratio of the probability of action for the action at time $t$ by the old agent before its weights were adjusted by PPO and the agent whose weights are being adjusted, that is, \(    ratio_{t}(\theta) = \frac{\pi(\bat|\bst,\theta)}{\pi_{\theta_{\text{old}}}(\bat|\bst)}\).
\hrule
\captionof{algorithm}{Proposed Algorithm}
\hrule
\begin{algorithmic}[1]
\Require SDP Solution of graph $G(V,E)$ given by $\mathbf{X}^{*} \in \mathbb{R}^{n\times n}$, Pretrained or untrained Policy \& Value Networks $\pi_{\theta_{old}}$ \&  $V_{\phi_{old}}$, learning rate $\lambda$, Buffer $Buf$ = \{\}, $\bso \sim \mathbb{U}(\mathbb{S}^{n-1})$, $t_{step}$, $n_{epochs}$
\Ensure Updated agent weights $\boldsymbol{\phi,\theta}$
\For{$t = 0,1,$\dots$,T$}
    \State $\bat \sim \pi_{\theta_{old}}(\bat|\bst) \quad \textit{(Sample action from old policy given current state)}$
    \State $\bss=\frac{\bat}{||\bat||_{2}}\quad\textit{(Transition to next state)}$
    \State $r_{t+1} = Cut(\bss)-Cut(\bst) \quad\textit{(Calculate reward)}$
    \State $V_{t} = r_{t+1} + V_{\phi_{\text{old}}}(\bss) \quad \textit{(Computing target for Value Network)}$
    \State $\hat{\delta}_{t,\pi_{old}} = V_{t} - V_{\phi_{old}}(\bst) \quad (\textit{Computing advantage estimate using one step TD})$
    \State $Buf \leftarrow Buf\cup \{ (\bst, \bat, \bss, V_{t}, \pi_{\theta_{old}}(\bat|\bst), \hat{\delta}_{t,\pi_{old}}) \}$
    \State $\bss \gets \bst$
    \If{$t \quad modulo \quad t_{step}=0$}
    \For {$epoch = 0,1,\dots,n_{epochs}$}
        \State Sample $(\bst, \bat, \bss, V_{t},\pi_{\theta_{old}}(\bat|\bst),\hat{\delta}_{t,\pi_{old}})$ from $Buf$
     \State Compute $L_{agent}$ as per \eqref{eqn:l_ppo}, \eqref{eqn:l_vf}, \eqref{eqn:l_agent}, 
    \State $\theta \leftarrow \theta - \lambda\nabla_{\theta}L_{agent}$
    \State $\phi \leftarrow \phi -\lambda \nabla_{\phi}L_{agent}$
    \EndFor
    \EndIf
    \State $\pi_{\theta_{old}}\leftarrow\pi_{\theta}$
    \State $V_{\phi_{old}}\leftarrow V_{\phi}$
\EndFor
\end{algorithmic}
\label{alg:gen_trajectory_update_agent}
\hrule

\section{Experiments}

\subsection{Settings}

We evaluate the agent on two graph datasets. The first consists of twelve 1000-node graphs from the well-known Gset dataset \cite{benson2000solving}. The second includes random 1000-node Erdős-Rényi (ER) graphs generated using the NetworkX package \cite{networkx}, with varying edge creation probabilities $p \in {0.1, 0.4, 0.8}$. For each value of $p$, we generate five graphs. We note that in ER graphs, $p$ controls the graph density; hence, we vary $p$ to evaluate our method across different densities\footnote{According to the three choices of $p$ in the ER graphs, the number of edges in these graphs approximately vary between $49950$ to $399600$ as the density of the graph is given as $\frac{2m}{n(n-1)}$. }. The choice of $n = 1000$ in both datasets reflects our aim to test on graphs that commercial ILP solvers struggle to scale with.

\begin{figure}[htbp]
    \centering
    
    \begin{subfigure}[t]{0.48\textwidth}
        \centering
        \includegraphics[width=\linewidth]{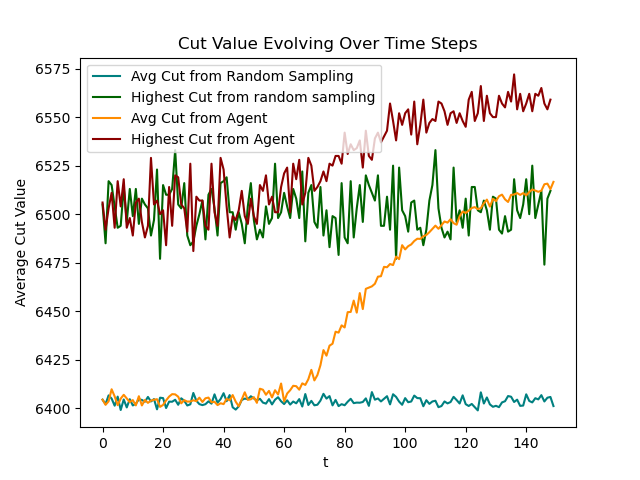}
        \caption{Average and maximum cut values for Gset graph seed $43$.}
        \label{fig:Gset43_main}
    \end{subfigure}
    \hfill
    \begin{subfigure}[t]{0.48\textwidth}
        \centering
        \includegraphics[width=\linewidth]{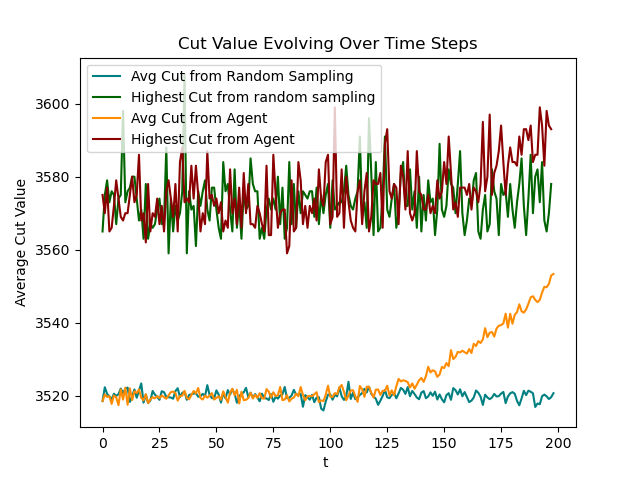}
        \caption{Average and maximum cut values for ER graph of seed $200$ and density $0.01$.
        }
        \label{fig:ER200_1000_main}
    \end{subfigure}
    
    \caption{Evolution of cut values (y-axis) over a batch of $256$ hyperplanes of our agent and pGW over time (divided by $10$ in the x-axis) for a Gset graph (a) and an ER graph (b), where ``random sampling'' refers to pGW.}
    \label{fig:comparison_gset_er}
\end{figure}

\begin{table*}[h]
\centering
\begin{tabular}{|c|cc|cc|}
\hline
\multicolumn{5}{|c|}{\textbf{Gset} graphs with $n=1000$} \\
\hline

\textbf{Gset instance} & \multicolumn{2}{c|}{\textbf{pGW}} & \multicolumn{2}{c|}{\textbf{Proposed Agent}}  \\
\cline{2-5}
 & \textbf{Avg} & \textbf{Max} & \textbf{Avg} & \textbf{Max} \\
\hline 
43 & 6404.9 & 6480 & 6546.6 & 6576 \\ \hline
44 & 6403.9 & 6487 & 6502.8 & 6535 \\ \hline
45 & 6402.9 & 6486 & 6477.1 & 6516 \\ \hline
46 & 6394.3 & 6465 & 6441.7 & 6486 \\ \hline
47 & 6408.3 & 6480 & 6481.1 & 6522 \\ \hline
51 & 3684.6 & 3739 & 3740.2 & 3759 \\ \hline
52 & 3689.1 & 3738 & 3735.5 & 3754 \\ \hline
53 & 3688.7 & 3740 & 3721.6 & 3752 \\ \hline
54 & 3686.3 & 3734 & 3740.6 & 3764 \\ \hline
\end{tabular}
\vspace{0.1in}
\caption{Average and maximum cut values for Gset graphs obtained from a batch of 256 hyperplanes of pGW against the proposed agent after 1500 time-steps. We see that the proposed agent is able to get a higher score than pGW on all seeds.}
\label{table:Gset_table_main}
\end{table*}

\begin{table*}[h]
\centering
\begin{tabular}{|c|c|cc|cc|}
\hline
\multicolumn{6}{|c|}{\textbf{Erdos-Renyi} graphs with $n=1000$} \\
\hline
\textbf{Seed}  &{$p$ (probability of edge creation)} & \multicolumn{2}{c|}{\textbf{pGW}} & \multicolumn{2}{c|}{\textbf{Proposed Agent}}  \\
\cline{3-6}
 &   & \textbf{Avg} & \textbf{Max} & \textbf{Avg}& \textbf{Max}\\
\hline 
1 & 0.1 & 28013.7 & 28181 & 28199 & 28292\\\hline

50 & 0.1 & 27874.3 & 28001 & 28069.7 & 28139\\\hline

100 & 0.1 & 27839.2 & 28015 & 28049.2 & 28131\\\hline

150 & 0.1 & 28000.7 & 28185 & 28173.9 & 28263\\\hline

200 & 0.1 & 27971.6 & 28180 & 28128.2 & 28209\\\hline

1 & 0.4 & 104689.8 & 105017 & 104944.3 & 105126\\\hline

50 & 0.4 & 104671.2 & 104970 & 104916.2 & 105070\\\hline

100 & 0.4 & 104952.2 & 105200 & 105186.5 & 105340\\\hline

150 & 0.4 & 105229.8 & 105557 & 105477.4 & 105599\\\hline

200 & 0.4 & 104941.2 & 105336 & 105161.8 & 105317\\\hline

1 & 0.8 & 203966.9 & 204260 & 204217.7 & 204404\\\hline

50 & 0.8 & 203796.1 & 204070 & 204066.8 & 204224\\\hline

100 & 0.8 & 204107.6 & 204407 & 204468.3 & 204607\\\hline

150 & 0.8 & 204264 & 204567 & 204518.5 & 204659\\\hline

200 & 0.8 & 203967.3 & 204299 & 203998.5 & 204284\\\hline

\end{tabular}
\vspace{0.1in}
\caption{Average and maximum cut values for the ER dataset obtained from a batch of 256 hyperplanes of pGW against the proposed Agent. All graphs here have a total of 1000 nodes and are ER graphs generated from NetworkX Package. We observe that the proposed agent is able to get a higher score than pGW on all seeds across all densities as indicated by the probability of edge creation (column 2). }
\label{table:ER_table_main}
\end{table*}

We begin with a batch of 256 normal vectors randomly sampled from the uniform distribution on the unit sphere. We emphasize that our approach does not rely on any training graphs, in contrast to typical ML- and RL-based MaxCut methods~\cite{s2v-dqn, eco-dqn, rl-bnb-2024}. Instead, we train and test directly on the same graph being optimized. The agent receives the SDP and proceeds to optimize the MDP for each graph using PPO.

For baselines, we use the pGW in Algorithm~\ref{alg:pgw} where we sample a batch of 256 unit norm vectors uniformly in parallel. We apply each of these vectors to~\eqref{eqn:reward} to obtain the value of Cut for a given graph. We report the average of these values as well as the maximum.

For our method, we use $l = 1024$ (number of hidden dimension) for our NN. See Appendix~\ref{sec: appen ablation} for an ablation study on $l$. 

We varied the learning rate and noticed that a learning rate of $0.001$ was sufficient to beat the score though better learning rates probably can lead to faster convergence above the average score and maximum score returned by the pGW algorithm. 

\subsection{Main Results}\noindent Here, we present results when using the setup described in the previous subsection. 
Figure~\ref{fig:Gset43_main} and Figure~\ref{fig:ER200_1000_main} show plots of the agent's average and maximum cut values among a batch of $256$ sampled hyperplanes while following the policy as compared to pGW for Gset seed 43 and the ER graph generated by NetworkX package with $1000$ nodes and density of $0.01$, respectively. 
Table~\ref{table:Gset_table_main} and Table~\ref{table:ER_table_main} present the complete results. As observed, in all cases, our agent achieves cut values that are higher than pGW, which is the main goal of this paper. 

\section{Discussion}
By leveraging reinforcement learning to adaptively select hyperplanes—rather than relying on uniform sampling as in the Goemans-Williamson (GW) algorithm—we demonstrate improved MaxCut values across various graph types. Notably, our agent is entirely self-supervised: it does not require any external training dataset and instead optimizes directly over the given graph instance. The approach is agnostic to graph structure, making it broadly applicable. While our method builds on the GW formulation by treating hyperplane selection as a policy optimization problem, we believe that the core framework—combining distributional action sampling with reward-based learning—can be extended to other combinatorial optimization problems. We leave such extensions to future work.

\bibliography{ArXiV}

\newpage
\appendix
\section{Impact of the number of parameters in our agent's NN} \label{sec: appen ablation}

Here, we include ablation studies on randomly generated ER graphs to determine if having higher number of learnable parameters in the encoder layer is beneficial to seeing higher cuts. To this end, we present results across different seeds, densities and number of hidden neurons in our NN, i.e., $l$. We ran ablation studies for 2000 time steps. 
We see that the average cut increased with increase in $l$ until $l=512$ and then decreased in Tables \ref{table:appendix_er1}, \ref{table:appendix_er3}, and, \ref{table:appendix_er5} whereas in \ref{table:appendix_er4} the average cut increased with higher values of $l$.

In the following tables, we present the average and maximum values for cut for our agent and compute the relative increase with respect to pGW. 

\begin{table*}[h]
\footnotesize
\centering
\begin{tabular}{|c|c|c|cc|cc|cc|}
\hline
\multicolumn{9}{|c|}{\textbf{ER}} \\
\hline
\textbf{Seed} &$p$ &$l$ &\multicolumn{2}{c|}{\textbf{pGW}} & \multicolumn{2}{c|}{\textbf{Proposed RL agent}}& \multicolumn{2}{c|}{\textbf{\% Increase}} \\
\cline{4-9}
 &  &  & \textbf{Avg} & \textbf{Max} & \textbf{Avg} & \textbf{Max}& \textbf{Avg} & \textbf{Max} \\
\hline 
1 & 0.29 & 64 & 77122.84 & 77369 & 77381.55 & 77552 &0.33&0.24\\ \hline
1 & 0.29 & 128 & 77122.84 & 77369& 77395.7 & 77539 &0.35&0.22\\ \hline
1 & 0.29 & 256 & 77122.84 & 77369 & 77397.83 & 77512 &0.36&0.18\\ \hline
1 & 0.29 & 512 & 77122.84 & 77369 & 77450.92 & 77556 &0.42&0.24\\ \hline
1 & 0.29 & 1024 & 77122.84 & 77369 & 77443.98 & 77542 &0.42&0.22\\ \hline
\end{tabular}
\vspace{0.1in}
\caption{Results of our agent using different number of hidden dimension, $l$ for ER graph with $p=0.29$ and seed $=1$. We see that the average cut generally increases until $l=512$.}
\label{table:appendix_er1}
\end{table*}

\begin{table*}[h]
\footnotesize
\centering
\begin{tabular}{|c|c|c|cc|cc|cc|}
\hline
\multicolumn{9}{|c|}{\textbf{ER}} \\
\hline
\textbf{Seed} & $p$ &$l$&\multicolumn{2}{c|}{\textbf{pGW}} & \multicolumn{2}{c|}{\textbf{Proposed RL agent}}& \multicolumn{2}{c|}{\textbf{\% Increase}} \\
\cline{4-9}
 &  &  & \textbf{Avg} & \textbf{Max} & \textbf{Avg} & \textbf{Max}& \textbf{Avg} & \textbf{Max} \\
\hline 
1 & 0.85 & 64 & 216019.09 & 216255 & 216020.92 & 216267 &0.001&0.07 \\ \hline
1 & 0.85 & 128 & 216019.09 & 216255  & 216167.16 & 216334 &0.07 &0.04\\ \hline
1 & 0.85 & 256 &216019.09 & 216255& 216185.1 & 216336 & 0.08&0.04\\ \hline
1 & 0.85 & 512 & 216019.09 & 216255& 216245.8 & 216358 &0.1&0.05\\ \hline
1 & 0.85 & 1024 &216019.09 & 216255& 216161.39 & 216345 &0.07&0.05\\ \hline
\end{tabular}
\vspace{0.1in}
\caption{Results of our agent using different number of hidden dimension, $l$ for ER graph with $p=0.85$ and seed $=1$. We see that the average cut increases until $l=512$ and then decreases.}
\label{table:appendix_er3}
\end{table*}
\begin{table*}[h]
\footnotesize
\centering
\begin{tabular}{|c|c|c|cc|cc|cc|}
\hline
\multicolumn{9}{|c|}{\textbf{ER}} \\
\hline
\textbf{Seed} & $p$ &$l$ &\multicolumn{2}{c|}{\textbf{pGW}} & \multicolumn{2}{c|}{\textbf{Proposed RL agent}}& \multicolumn{2}{c|}{\textbf{\% Increase}} \\
\cline{4-9}
 &  &  & \textbf{Avg} & \textbf{Max} & \textbf{Avg} & \textbf{Max}&\textbf{Avg} & \textbf{Max} \\
\hline 
50 & 0.29 & 64 & 76820.31 & 77111 & 77039.88 & 77267&0.29&0.20  \\ \hline
50 & 0.29 & 128 & 76820.31 & 77111   & 77118.55 & 77300 &0.39&0.24 \\ \hline
50 & 0.29 & 256 &76820.31 & 77111 & 77111.06&77272 &0.38&0.21\\ \hline
50 & 0.29 & 512 & 76820.31 & 77111 & 77174.1 & 77290 &0.46&0.23\\ \hline
50 & 0.29 & 1024 &76820.31 & 77111 & 77185.7 & 77299 &0.48&0.24\\ \hline
\end{tabular}
\vspace{0.1in}
\caption{Results of our agent using different number of hidden dimension, $l$ for ER graph with $p=0.29$ and seed $=50$. We see that the average cut from agent have an increasing trend with increase in hidden dimension.}
\label{table:appendix_er4}
\end{table*}
\begin{table*}[h]
\centering
\begin{tabular}{|c|c|c|cc|cc|cc|}
\hline
\multicolumn{9}{|c|}{\textbf{ER}} \\
\hline
\textbf{Seed} & $p$ &$l$&\multicolumn{2}{c|}{\textbf{pGW}} & \multicolumn{2}{c|}{\textbf{Proposed RL agent}}& \multicolumn{2}{c|}{\textbf{\% Increase}} \\
\cline{4-9}
 &  &  & \textbf{Avg} & \textbf{Max} & \textbf{Avg} & \textbf{Max}& \textbf{Avg} & \textbf{Max} \\
\hline 
50 & 0.85 & 64 & 215870.01 & 216146 & 215894.8 & 216193   &0.01&0.02 \\  \hline
50 & 0.85 & 128 & 215870.01 & 216146  & 215925.86 & 216190 &0.03&0.02 \\ \hline
50 & 0.85 & 256 &215870.01 & 216146& 216021.88 & 216202 &0.07&0.03\\ \hline
50 & 0.85 & 512 & 215870.01 & 216146& 216096.23 & 216236&0.1&0.04 \\ \hline
50 & 0.85 & 1024 &215870.01 & 216146& 216056.19 & 216269 &0.09&0.06\\ \hline
\end{tabular}
\vspace{0.1in}
\caption{Results of our agent using different number of hidden dimension, $l$ for ER graph with $p=0.85$ and seed $=50$. We see that the average cut from agent have an increasing trend with increase in hidden dimension until $l=512$ after which it decreases slightly.}
\label{table:appendix_er5}
\end{table*}

\end{document}